\documentclass[10pt,twocolumn,letterpaper]{article}

\usepackage{wacv}
\usepackage{times}
\usepackage{epsfig}
\usepackage{graphicx}
\usepackage{amsmath}
\usepackage{amssymb}

\usepackage{booktabs}
\usepackage{caption}
\usepackage{dsfont}
\usepackage{graphicx}
\usepackage{tabularx}
\usepackage{adjustbox}
\newcolumntype{L}[1]{>{\raggedright\arraybackslash}p{#1}}
\newcolumntype{C}[1]{>{\centering\arraybackslash}p{#1}}
\newcolumntype{R}[1]{>{\raggedleft\arraybackslash}p{#1}}
\newcommand{\varx}{\mathbf{\mspace{1mu}x}}
\newcommand{\vary}{\mathbf{\mspace{1mu}y}}
\newcommand{\varz}{\mathbf{\mspace{1mu}z}}

\DeclareMathOperator*{\argmax}{argmax}

\DeclareMathOperator*{\softmax}{softmax}
\usepackage{bbm}
\usepackage{enumitem}
\usepackage{multirow}
\setlist[enumerate,1]{label=\arabic*}
\newlist{inlinelist}{enumerate*}{1}
\setlist*[inlinelist,1]{label=(\roman*)}

\def\wacvPaperID{775} 

\wacvfinalcopy 
\ifwacvfinal
\def\assignedStartPage{9876} 
\fi

\ifwacvfinal
\usepackage[breaklinks=true,bookmarks=false]{hyperref}
\else
\usepackage[pagebackref=true,breaklinks=true,colorlinks,bookmarks=false]{hyperref}
\fi

\ifwacvfinal
\else
\fi

\begin{document}

\title{Integrating Human Gaze into Attention for Egocentric Activity Recognition}

\author{Kyle Min\quad Jason J. Corso\\
University of Michigan\\
Ann Arbor, MI 48109\\
{\tt\small \{kylemin,jjcorso\}@umich.edu}
}

\maketitle

\begin{abstract}
It is well known that human gaze carries significant information about visual attention. However, there are three main difficulties in incorporating the gaze data in an attention mechanism of deep neural networks:
\begin{inlinelist}
  \item the gaze fixation points are likely to have measurement errors due to blinking and rapid eye movements;
  \item it is unclear when and how much the gaze data is correlated with visual attention; and
  \item gaze data is not always available in many real-world situations.
\end{inlinelist}
In this work, we introduce an effective probabilistic approach to integrate human gaze into spatiotemporal attention for egocentric activity recognition. Specifically, we represent the locations of gaze fixation points as structured discrete latent variables to model their uncertainties. In addition, we model the distribution of gaze fixations using a variational method. The gaze distribution is learned during the training process so that the ground-truth annotations of gaze locations are no longer needed in testing situations since they are predicted from the learned gaze distribution. The predicted gaze locations are used to provide informative attentional cues to improve the recognition performance. Our method outperforms all the previous state-of-the-art approaches on EGTEA, which is a large-scale dataset for egocentric activity recognition provided with gaze measurements. We also perform an ablation study and qualitative analysis to demonstrate that our attention mechanism is effective.
\end{abstract}

\section{Introduction} \label{sec:intro}
It has recently been shown that attention mechanisms can boost the performance of neural networks in various tasks by learning to focus on relatively important and salient parts of input signals. Most notably, attention-based recurrent neural networks have achieved great success in machine translation~\cite{bahdanau2014neural, luong2015effective} and image captioning~\cite{xu2015show}. Attention mechanisms have also been widely adopted by deep convolutional neural networks (CNNs) in several forms of feature re-weighting such as spatial attention~\cite{xu2016ask, oktay2018attention}, channel attention~\cite{hu2018squeeze, zhang2018image}, etc~\cite{woo2018cbam, suganuma2018attention}. These methods usually let neural networks learn \textit{what and where} to focus on from their own responses.

In this paper, we introduce an effective probabilistic method for integrating human gaze into a spatiotemporal attention mechanism. It has been well discussed in cognitive science that human gaze is closely related to a person's behavioral intention and visual attention~\cite{vickers2009advances, castiello2003understanding, frischen2007gaze, phillips2002infants}. At the same time, however, there is always uncertainty in the process of recording the gaze fixation points because of saccadic suppression\footnote{phenomenon in which visual information is not processed while blinking or under rapid eye movements.}\cite{krekelberg2010saccadic} and measurement errors. Furthermore, it is not always guaranteed that the surrounding region around the point of gaze fixation has the most important information, especially when interacting with multiple objects or under dissociation\footnote{dissociation of the focus of attention is a phenomenon where the points of gaze fixation are not correlated with the visual attention within the field of view.}\cite{brefczynski1999physiological, juan2004dissociation}.

To address such problems, we present a probabilistic modeling method as follows: First, we propose to represent the locations of gaze fixation points in space and time as structured discrete latent variables to model their uncertainties. Second, we model the distribution of the gaze fixations using a variational method. During the training process, the distribution of gaze fixations is learned using the ground-truth annotations of gaze points. Specifically, we propose to reformulate the discrete training objective so that it can be optimized using an unbiased gradient estimator. The gaze locations are predicted from the learned gaze distribution so that the ground-truth annotations of gaze fixation points are no longer needed in testing scenarios. The predicted gaze locations are integrated into a soft attention mechanism to make the intermediate features more attended to informative regions. It is empirically shown that our gaze-combined attention mechanism leads to a significant improvement of activity recognition performance on egocentric videos by providing additional cues across space and time.

We demonstrate the effectiveness of our method on EGTEA~\cite{li2018eye} and GTEA gaze+~\cite{li2015delving}, which are large-scale datasets for egocentric activities provided with gaze measurements. Our method significantly outperforms all the previous state-of-the-art approaches. We also perform an ablation study to verify that probabilistic modeling of gaze data is truly beneficial. We then visualize the spatiotemporal responses of our networks to qualitatively show that the gaze-combined soft attention provides informative attentional cues.

\section{Related work} \label{sec:related}
Recently, attention-based recurrent neural networks have been widely adopted for neural machine translation~\cite{bahdanau2014neural,luong2015effective} as well as for image captioning~\cite{xu2015show}. They generate attention vectors by manipulating hidden states of recurrent neural networks and annotated information. Attention mechanisms have also been incorporated with deep CNNs to improve the representation quality of intermediate features by refining the features~\cite{xu2016ask,oktay2018attention,hu2018squeeze,zhang2018image}. They usually introduce attention modules which find channel-wise or spatial-wise attention maps from the average-pooled features descriptors. There are more recent works which utilize both attention methods across spatial and channel dimensions~\cite{woo2018cbam,suganuma2018attention}. These methods also have shown that using both average-pooling and max-pooling in parallel is beneficial to building attention maps.

There have been a few attempts to utilize human gaze data for egocentric activity recognition~\cite{fathi2012learning,huang2019mutual,li2018eye}. Fathi~et~al.~\cite{fathi2012learning} propose a conditional generative model that jointly predicts gaze locations and egocentric activity labels. More related and recent works~\cite{huang2019mutual,li2018eye} have shown that incorporating gaze data into an attention mechanism can boost the performance of CNNs on egocentric activity recognition. Huang~et~al.~\cite{huang2019mutual} propose Mutual Context Network (MCN) that tries to use human gaze for recognizing activities and use the activity labels for predicting gaze locations. However, MCN has multiple sub-modules that should be trained separately. Furthermore, an inference procedure requires many iterations because of the complicated network architecture. They also use saccades as ground-truth gaze points, which should be ignored to improve the prediction performance. Li~et~al.~\cite{li2018eye} is built on a similar probabilistic framework to ours; however, there are three crucial differences. First, to model the distribution of gaze points for $T$ time steps, they use $T$ independent 2D latent variables. This totally ignores the temporal correlation of the gaze distribution, which limits the recognition performance. Second, they use the approximated Gumbel-Softmax objective~\cite{jang2016categorical,maddison2016concrete} that introduces a significant bias to a gradient estimator. As a result, the recognition performance of their method is further limited. Third, they directly apply the sampled gaze points $\varz^*$ to the input feature map without any modifications. This is vulnerable to situations where the gaze points are misleading and not informative. On the contrary, we use structured discrete latent variables to model the gaze distribution in a 3D space. We apply the direct optimization method to handle this structured latent space, which also minimizes the bias. Moreover, we use the sigmoid activated linear mapping on the sampled gaze points to produce a soft attention map.

\section{Background: Direct optimization} \label{sec:background}
Direct optimization~\cite{lorberbom2018direct} was originally proposed for learning a variational auto-encoder (VAE) with discrete latent variables. The objective of VAE is given by:
\begin{equation}
\label{eq:vae-loss}
\mathcal{L}_{\text{VAE}}=-\mathbb{E}_{\varz \sim q_{\phi}}[\log p_{\theta}(\varx|\varz)] + D_{\mathrm{KL}}[q_{\phi}(\varz|\varx)||p_{\theta}(\varz)]
\end{equation}
where $\varx$ is an input and $\varz$ is a discrete latent variable. Computing the expected log-likelihood requires drawing samples from the discrete distribution $q_{\phi}(\varz|\varx)$, which makes it difficult to optimize. Gumbel-Softmax reparameterization technique~\cite{jang2016categorical,maddison2016concrete} was recently suggested to relax the discrete variables to continuous counterparts. However, this continuous relaxation is known to introduce a significant bias when evaluating gradients and become intractable under the high-dimensional structured latent spaces. The direct optimization method introduces an unbiased gradient estimator for the discrete VAE that can be used even under the high-dimensional structured latent spaces. For simplicity, let us rewrite the log-probabilities as follows: $h_{\phi}(\varx, \varz)=\log q_{\phi}(\varz|\varx)$, $f_{\theta}(\varx, \varz)=\log p_{\theta}(\varx|\varz)$. By using the Gumbel-Max trick~\cite{maddison2014sampling}, the expected log-likelihood can be reformulated as follows: $\mathbb{E}_{\varz \sim q_{\phi}}[\log p_{\theta}(\varx|\varz)] = \mathbb{E}_{\gamma \sim G}[f_{\theta}(\varx,\varz^*)]$ where $\varz^*=\argmax_{\hat{\varz}} \{ h_{\phi}(\varx,\hat{\varz}) +\gamma(\hat{\varz})\}$, G denotes a Gumbel distribution, and $\gamma(\hat{\varz})$ represents a random variable sampled from the Gumbel distribution that is associated with each input $\hat{\varz}$. Then, the proposed gradient estimator for the expectation term is given in the following form:
\begin{multline}
\label{eq:vae-direct}
\nabla_{\phi} \, \mathbb{E}_{\gamma \sim G}[f_{\theta}(\varx,\varz^*)] =  \lim\limits_{\epsilon \to 0} \frac{1}{\epsilon} \Big( \mathbb{E}_{\gamma \sim G}\big[\nabla_{\phi} \, h_{\phi} (\varx,\varz^*(\epsilon))
\\ - \nabla_{\phi} \, h_{\phi}(\varx,\varz^*) \big] \Big)
\end{multline}
where $\varz^*(\epsilon)=\argmax_{\hat{\varz}} \{ \epsilon f_{\theta}(\varx,\hat{\varz}) +  h_{\phi}(\varx,\hat{\varz}) +\gamma(\hat{\varz})\}$. The suggested gradient estimator is unbiased when the perturbation parameter $\epsilon$ goes to 0, but small $\epsilon$ brings a large variance of the estimation. Therefore, in practice, we set $\epsilon$ to a large value in the beginning of the training process and decrease it progressively.

\section{Method} \label{sec:method}
We start this section by building a probabilistic framework and the loss function of our method. Next, we propose a 3D gaze modeling approach using structured discrete latent variables. We then introduce the direct loss minimization approach~\cite{lorberbom2018direct} that is used for optimization in the presence of the structured discrete latent variables. Finally, we describe our overall network architecture for activity recognition that integrates the gaze information into attention.

\subsection{Probabilistic framework} \label{subsec:framework}
Let us consider a recognition task of predicting activity labels $\vary$ given an input clip of egocentric videos $\varx$, which is equivalent to finding a conditional probability $p(\vary|\varx)$. We represent the gaze locations in space and time with a discrete latent variable $\varz$. Then, the conditional probability is written as follows by the law of total probability:
\begin{equation}
\label{eq:pf1}
p_{\theta}(\vary|\varx) = \int p_{\theta}(\vary|\varx,\varz)p_{\theta}(\varz|\varx)d\varz
\end{equation}
where $\theta$ denotes the parameters of a network for recognition. Since $\varz$ generally has an intractable posterior distribution, we upper bound the negative log-likelihood by taking the negative log on both sides of Equation~(\ref{eq:pf1}) and introducing the variational approximation $q_{\phi}(\varz|\varx)$ for gaze modeling as follows:
\begin{multline}
\label{eq:pf2}
-\log p_{\theta}(\vary|\varx)
\leq \int -q_{\phi} \log \Big(p_{\theta}(\vary|\varx,\varz)\frac{p_{\theta}(\varz|\varx)}{q_{\phi}} \Big) d\varz \\
= -\mathbb{E}_{\varz \sim q_{\phi}}[\log p_{\theta}(\vary|\varx,\varz)] + D_{\mathrm{KL}}[q_{\phi}||p_{\theta}(\varz|\varx)]
\end{multline}
where $\phi$ denotes parameters of a network for gaze modeling. We use the upper bound in Equation~(\ref{eq:pf2}) as our loss function.

\subsection{Reformulating the training objective} \label{subsec:objective}
In order to compute the expected log-likelihood of the loss function in Equation~(\ref{eq:pf2}), we need to sample the gaze points from $q_{\phi}$. We apply the Gumbel-Max trick~\cite{maddison2014sampling} that is an efficient method of drawing samples from a discrete distribution. For simplicity, let us rewrite the log-probability as follows: $h_{\phi}(\varx, \varz)=\log q_{\phi}(\varz|\varx)$. Then, we can draw a gaze sample $z^*$ using the following equation:
\begin{equation}
\label{eq:opt1}
\varz^*=\argmax \limits_{\hat{\varz}} \{ h_{\phi}(\varx,\hat{\varz}) +\gamma(\hat{\varz})\}
\end{equation}
where $\gamma(\hat{\varz})$ represents a random variable sampled from a Gumbel distribution that is associated with each input $\hat{\varz}$. However, $\varz^*$ includes a non-differentiable operation, $\argmax$, so we cannot evaluate the gradient of the expectation term with respect to $\phi$ using a standard backpropagation algorithm. Here, we propose to apply the direct optimization method~\cite{lorberbom2018direct} to optimize the expected log-likelihood term. In the following, we demonstrate that our loss function can be optimized using the direct optimization method.

Since our task is to classify activity labels, we can model $\vary$ given $\varx$ and $\varz$ with a categorical distribution. Specifically, let us say that there are $C$ number of predefined activity classes. Then, $p_{\theta}(\vary|\varx,\varz)=\prod_{c=1}^{C}p_{c}^{\mathbbm{1}_{y=c}}$ for some class-wise probabilities $p_{c}$'s that are dependent on $\varx$ and $\varz$ where $\mathbbm{1}_{y=c}$ is an indicator function that is equal to 1 if $y=c$ and 0 otherwise. This allows us to rewrite $\log p_{\theta}(\vary|\varx,\varz)$ in the following form:
\begin{equation}
\label{eq:opt2}
\log p_{\theta}(\vary|\varx,\varz)=\sum_{c=1}^{C}\mathbbm{1}_{y=c} \, f_{\theta}^{c}(\varx,\varz)
\end{equation}
where $f_{\theta}^{c}(\varx,\varz)$'s are the corresponding class-wise log-probabilities. Now, we propose to reformulate the expected log-likelihood using the class-wise log-probabilities:
\begin{align}
\mathbb{E}_{\varz \sim q_{\phi}}[\log p_{\theta}] 
&= \sum_{\varz} \Big( \mathbb{P}_{\gamma \sim G}[\varz^* = \varz] \sum_{c=1}^{C}\mathbbm{1}_{y=c} \, f_{\theta}^{c}(\varx,\varz) \Big) \nonumber \\
\label{eq:opt3}
&= \sum_{c=1}^{C} \mathbbm{1}_{y=c} \, \mathbb{E}_{\gamma \sim G}[f_{\theta}^{c}(\varx,\varz^*)]
\end{align}
where $G$ denotes the Gumbel distribution. In Equation~(\ref{eq:opt3}), We show that the expected log-likelihood can be decomposed into a sum of multiple expectation terms of the class-wise log-probabilities, each multiplied by an indicator function. Since the gradient is a linear operator, we can estimate the gradient of the expected log-likelihood as follows:
\begin{equation}
\label{eq:opt4}
\nabla_{\phi} \, \mathbb{E}_{\varz \sim q_{\phi}}[\log p_{\theta}] 
= \sum_{c=1}^{C} \mathbbm{1}_{y=c} \, \nabla_{\phi} \, \mathbb{E}_{\gamma \sim G}[f_{\theta}^{c}(\varx,\varz^*)]
\end{equation}
where each class-wise gradient estimator $\nabla_{\phi} \, \mathbb{E}_{\gamma \sim G}[f_{\theta}^{c}(\varx,\varz^*)]$ is computed by applying the direct optimization:
\begin{multline}
\label{eq:opt5}
\nabla_{\phi} \, \mathbb{E}_{\gamma \sim G}[f_{\theta}^{c}(\varx,\varz^*)] = \lim\limits_{\epsilon \to 0} \frac{1}{\epsilon} \Big( \mathbb{E}_{\gamma \sim G}\big[\nabla_{\phi} \, h_{\phi} (\varx,\varz^*(\epsilon,c))
\\ - \nabla_{\phi} \, h_{\phi}(\varx,\varz^*) \big] \Big)
\end{multline}
when $\varz^*(\epsilon,c)=\argmax_{\hat{\varz}} \{ \epsilon f_{\theta}^{c}(\varx,\hat{\varz}) +  h_{\phi}(\varx,\hat{\varz}) +\gamma(\hat{\varz})\}$. Other gradients, such as the gradient of the expected log-likelihood with respect to $\theta$, are obtained using a standard backpropagation algorithm. As a result of the reformulation, we can optimize the training objective without introducing a bias of gradient estimator.

\subsection{Structured gaze modeling} \label{subsec:gaze-modeling}

\begin{figure*}[t]
  \centering
  \includegraphics[width=0.9\linewidth]{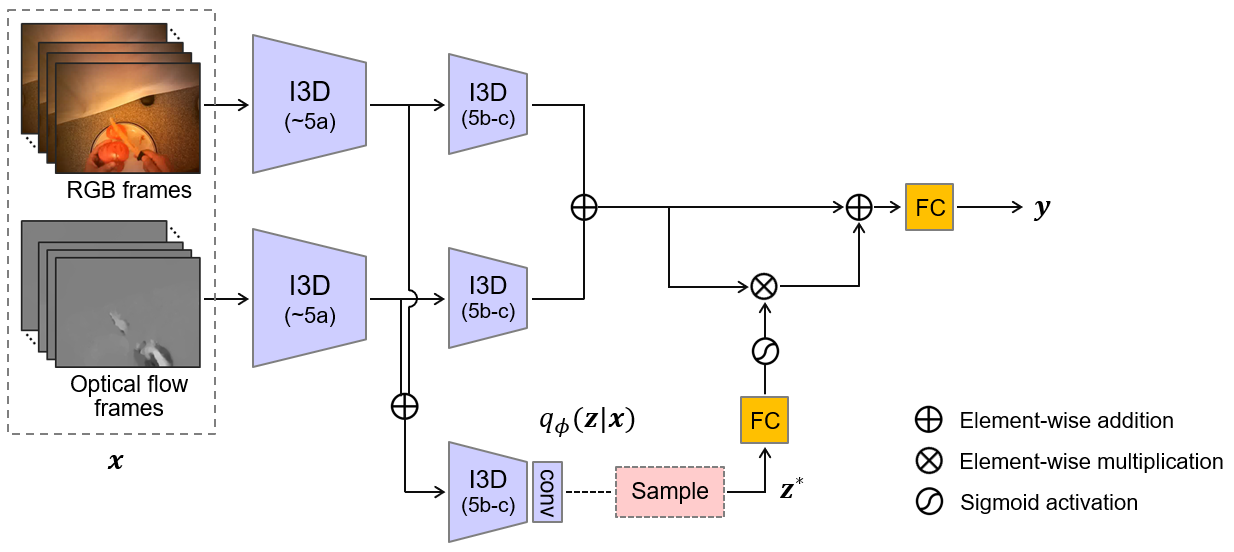}
  \caption{An illustration of our overall network architecture. We use the two-stream I3D~\cite{carreira2017quo} as a backbone network. To model the gaze distribution $q_{\phi}(\varz|\varx)$, we use the same convolutional blocks of the I3D (\texttt{Mixed\_5b-c}) and add three convolutional layers (conv) on top of it. The two intermediate features at the end of the 4th max-pooling layer (\texttt{MaxPool\_5a}) are added in an element-wise fashion and used as input to the network for gaze modeling. The sampled gaze point is applied with a fully-connected layer (FC) and with the sigmoid function to produce a soft attention map.
  \label{fig:network}
  }
\end{figure*}

We propose to use structured discrete latent variables to model the gaze locations as follows. First, we will write $\mathcal{Z}$ to denote a set of every possible $\varz$. Let us say that we want to model the gaze locations in a 3D space: $\mathcal{Z}=\mathbb{R}^{T \times H \times W}$ where $T$ is the length of the temporal dimension and $H$ and $W$ represent the height and width of spatial dimensions. For each time step, gaze is fixated at a single location of a $H \times W$ dimensional space. Therefore, it is more reasonable to represent the gaze locations with a sequence of 2D discrete random variables rather than with a single 3D random variable. Specifically, we assign a 2D discrete random variable to each time step: $\varz=(\varz_1,...,\varz_t,...,\varz_T)$ where each $\varz_t$ is one-hot encoded. For example, if the gaze is fixated at $(h,w)$ on the $t$-th time step, $\varz_{t}(j,k)=1$ if $(j,k)=(h,w)$ and 0 otherwise.

Computing $\varz^*(\epsilon,c)$ in Equation~(\ref{eq:opt5}) requires evaluating $f_{\theta}^{c}(\varx,\varz)$ for every $\varz$, which causes serious overhead. Although our structured gaze modeling reduces the number of possible realizations from $2^{THW}$ to $(HW)^{T}$, it is still computationally expensive. We propose to further reduce the number of computations by applying a low-dimensional approximation as suggested by Lorberbom et al.~\cite{lorberbom2018direct}. In particular, we approximate $f_{\theta}^{c}(\varx,\varz)=\sum_{t=1}^{T} f_{t}^{c}(\varx,\varz_{t};\theta)$ where $f_{t}^{c}(\varx,\varz_{t};\theta)=f_{\theta}^{c}(\varx,\varz_{1}^{*},...,\varz_{t},...\varz_{T}^{*})$. This low-dimensional approximation further reduces the number of possible realizations from $(HW)^{T}$ to $THW$. We implement the realization of $\varz$ by using the batch operation so that we can obtain $\varz^*(\epsilon,c)$ in a single forward pass.

\subsection{Network architecture} \label{subsec:network}

The overall network architecture is illustrated in Figure~\ref{fig:network}. As a backbone network, we use the two-stream I3D~\cite{carreira2017quo} which is a popular network for activity recognition tasks (\#Params: 24.7M, FLOPs: 80.2G). To model the gaze distribution $q_{\phi}(\varz|\varx)$, we use the same convolutional blocks of the I3D (\texttt{Mixed\_5b-c}) and add three convolutional layers (kernel size=[(1,3,3), (1,3,3), (1,1,1)], stride=[(1,1,1), (1,1,1), (1,1,1)]) on top of it. We add the two intermediate features at the end of the 4th max-pooling layer (\texttt{MaxPool\_5a}) and use the added feature map as an input to the network for gaze modeling. We draw a sample $\varz^*$ using the Equation~(\ref{eq:opt1}), which is then applied with a fully connect layer and the sigmoid function to produce a soft attention map. The two features at the end of the 5th convolutional block (\texttt{Mixed\_5c}) are added in an element-wise way, and we apply the soft attention map to the added feature map via a residual connection. Our final network has \#Params: 31.9M, FLOPs: 81.3G.

\section{Experiments} \label{sec:exp}
We evaluate our method on EGTEA~\cite{li2018eye}, which is a large-scale dataset with over 10k video clips of 106 fine-grained egocentric activities and annotated gaze fixations. It is demonstrated that our method outperforms other previous state-of-the-art approaches. Furthermore, we provide a qualitative analysis by visualizing the spatiotemporal responses of our network. We perform additional experiments on GTEA Gaze+~\cite{li2015delving} that consists of 2k videos with 44 activity categories.

\begin{table*}[t]
\centering
\begin{tabular}{L{4.6cm}|L{4.6cm}|C{2.3cm}|C{2.3cm}}
\toprule
Method & Backbone network & Acc (\%) & Acc$^{*}$ (\%) \\ \midrule
Li~et~al.~\cite{li2018eye} & I3D~\cite{carreira2017quo} & 53.30 & - \\
Sudhakaran~et~al.~\cite{sudhakaran2018attention} & ResNet34+LSTM~\cite{he2016deep,xingjian2015convolutional} & - & 60.76 \\
LSTA~\cite{sudhakaran2019lsta} & ResNet34+LSTM~\cite{he2016deep,xingjian2015convolutional} & - & 61.86 \\
MCN~\cite{huang2019mutual} & I3D~\cite{carreira2017quo} & 55.63 & - \\
Kapidis~et~al.~\cite{kapidis2019multitask} & MFNet~\cite{chen2018multi} & 59.44 & 66.59 \\
Lu~et~al.~\cite{lu2019learning} & I3D~\cite{carreira2017quo} & 60.54 & 68.60 \\ \midrule
Ours & I3D~\cite{carreira2017quo} & \textbf{62.84} & \textbf{69.58} \\
\bottomrule
\end{tabular}
\caption{Performance comparison of our method with other state-of-the-art methods on EGTEA dataset~\cite{li2018eye}. We report both Acc (mean class accuracy) and Acc$^{*}$ (ratio of correctly classified videos to the total number of videos). Acc is typically lower than Acc$^{*}$ due to an imbalanced class distribution of the dataset.}
\label{tab:act}
\end{table*}

\begin{table*}[t]
\centering
\begin{tabular}{L{4.6cm}|L{4.6cm}|C{2.3cm}|C{2.3cm}}
\toprule
Method & Backbone network & Acc (\%) & Acc$^{*}$ (\%) \\ \midrule
Sudhakaran~et~al.~\cite{sudhakaran2018attention} & ResNet34+LSTM~\cite{he2016deep,xingjian2015convolutional} & - & 60.13 \\
MCN~\cite{huang2019mutual} & I3D~\cite{carreira2017quo} & 61.14 & - \\
Ma~\textit{et~al.}~\cite{ma2016going} & FCN32s+CNN-M-2048~\cite{long2015fully,chatfield2014return} & - & 66.40 \\
Shen~\textit{et~al.}~\cite{shen2018egocentric} & SSD+LSTM~\cite{liu2016ssd} & - & 67.10 \\
\midrule
Ours & I3D~\cite{carreira2017quo} & \textbf{64.81} & \textbf{68.67} \\
\bottomrule
\end{tabular}
\caption{Performance comparison on the GTEA Gaze+~\cite{li2015delving} dataset. We report both Acc (mean class accuracy) and Acc$^{*}$ (ratio of correctly classified videos to the total number of videos). Ours again achieves the best performance.}
\label{tab:gtea}
\end{table*}

\subsection{Implementation details} \label{subsec:imp}
\textbf{Training/testing process. } First, we resize each frame to $256\times340$ and generate optical flow frames by using the TV-L1 algorithm~\cite{zach2007duality}. Following the previous works on the EGTEA dataset~\cite{huang2019mutual,li2018eye}, we use the I3D pre-trained on Kinetics dataset~\cite{carreira2017quo} as a backbone network. During the training process, we randomly sample 24-frame input segments and randomly crop $224\times224$ regions for each segment. We train our network in an end-to-end manner with a batch size of 24 on 8299 training video clips using the first split of the dataset. We use the SGD algorithm with 0.9 momentum and 0.00004 weight decay. The learning rate starts at 0.032 and decays two times by a factor of 10 after 8k and 15k iterations. $\epsilon$ is set to 1000 in the beginning and decreases exponentially with a 0.001 annealing rate. We set the minimal $\epsilon$ to be 0.1. $\epsilon$ goes to this minimum value within 10k iterations. The whole training process of 18K iterations takes less than 12 hours using 4 GPUs (TITAN Xp). For the evaluation, we divide each testing video into non-overlapping 24-frame segments. The whole evaluation process takes less than a half hour using a single GPU.

\textbf{Dimensions of the latent space. } For better comparison, we decided to follow the previous approaches for the dimensions of the latent space. Li~et~al.~\cite{li2018eye} suggests predicting gaze points for every 8 frames using the fact that a common duration of gaze fixation is roughly the same as the time interval of 8 frames (about 300ms). It is also suggested to reduce the spatial dimensions of the space for gaze distribution by a factor of 32. This is reasonable since our final goal is to improve the recognition performance, not to predict the exact gaze location in a high-dimensional space. As a result, the dimensions of the 3D latent space for gaze points described in Section~\ref{subsec:gaze-modeling} become $\mathcal{Z}=\mathbb{R}^{3 \times 7 \times 7}$ as $T=24/8$ and $H=W=224/32$.

\subsection{Comparison with the State-of-the-art} \label{subsec:comparison}

We compare our method with other state-of-the-art methods. Performance comparison on the EGTEA dataset is reported in Table~\ref{tab:act}. We want to point out that Li~et~al.~\cite{li2018eye}, MCN~\cite{huang2019mutual}, and Lu~et~al.~\cite{lu2019learning} use the same backbone network as ours, which is two-stream I3D~\cite{carreira2017quo}. Our method outperforms all other methods by a large margin.

We also evaluate our method on the GTEA Gaze+~\cite{li2015delving}, which is another commonly-used dataset for egocentric activity recognition provided with gaze measurements. It is collected by 6 different human subjects. Following previous works, we perform a leave-one-subject-out cross validation. The performance comparison is reported in Table~\ref{tab:gtea}. Our method again achieves the best performance among the recent approaches.

\begin{figure*}[t]
\centering
\hspace{-6pt}
  \adjustbox{valign=t}{\begin{minipage}[t]{0.49\textwidth}
  \small
    \includegraphics[width=1\textwidth]{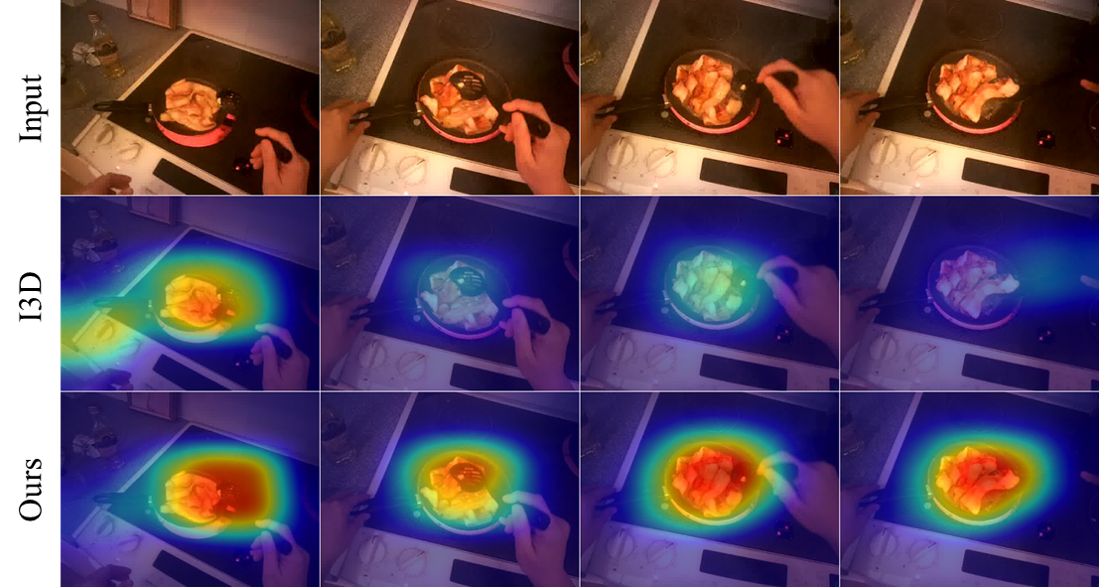}\\[-0.5ex] \hspace*{13.7em}(a)\\[0.7ex]
    \includegraphics[width=1\textwidth]{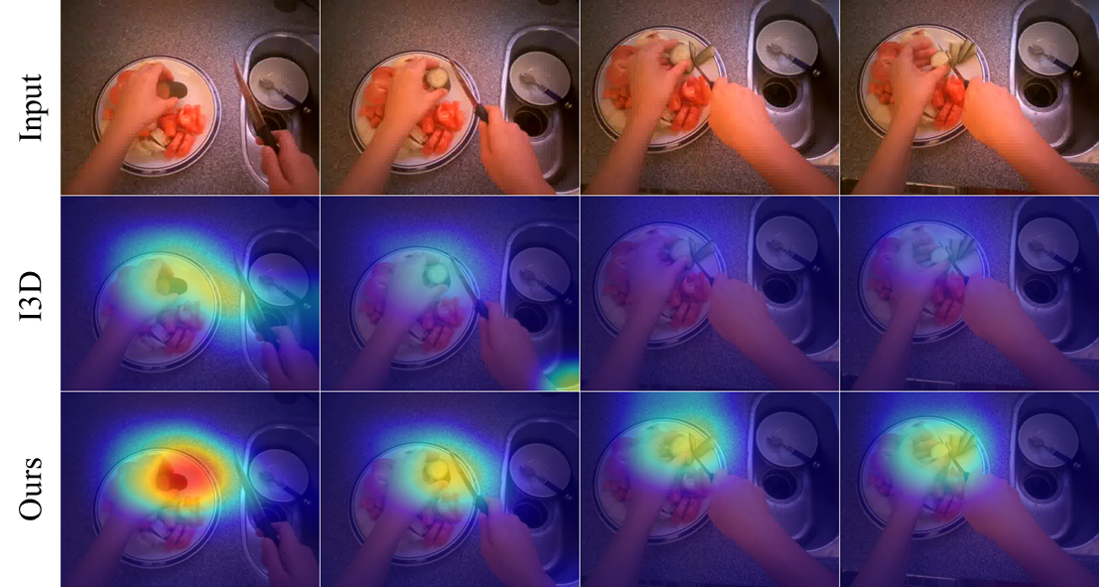}\\[-0.5ex] \hspace*{13.7em}(b)\\[-0.8ex]
    \hspace*{1.2em}\includegraphics[width=0.925\textwidth]{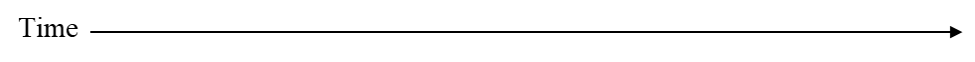}
  \end{minipage}}
  \hspace{0.029cm}\vline\hspace{0.02cm}
  \adjustbox{valign=t}{\begin{minipage}[t]{0.49\textwidth}
  \small
    \includegraphics[width=1\textwidth]{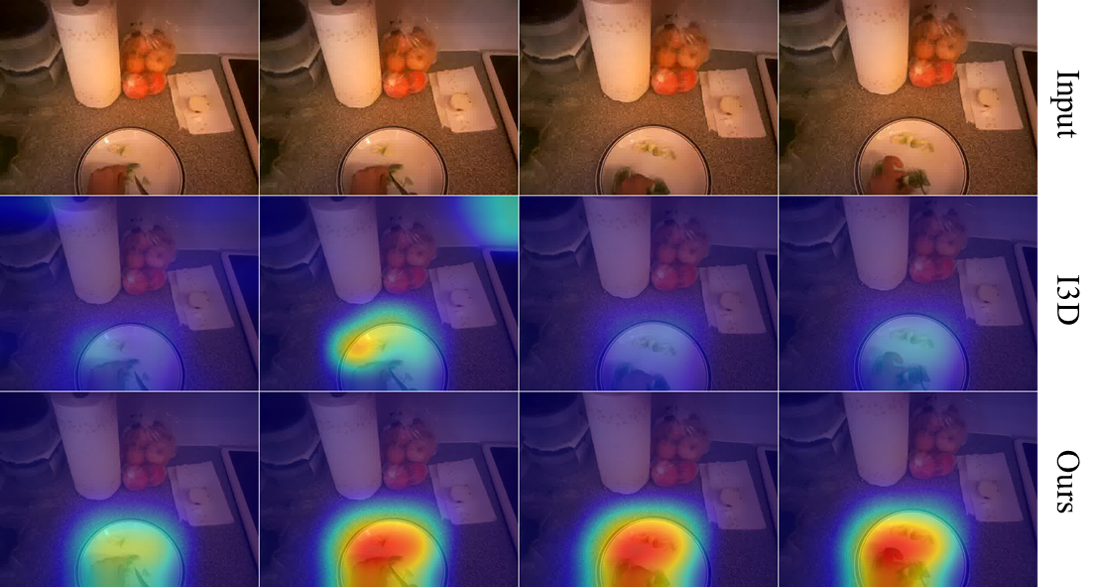}\\[-0.5ex] \hspace*{12.2em}(c)\\[0.7ex]
    \includegraphics[width=1\textwidth]{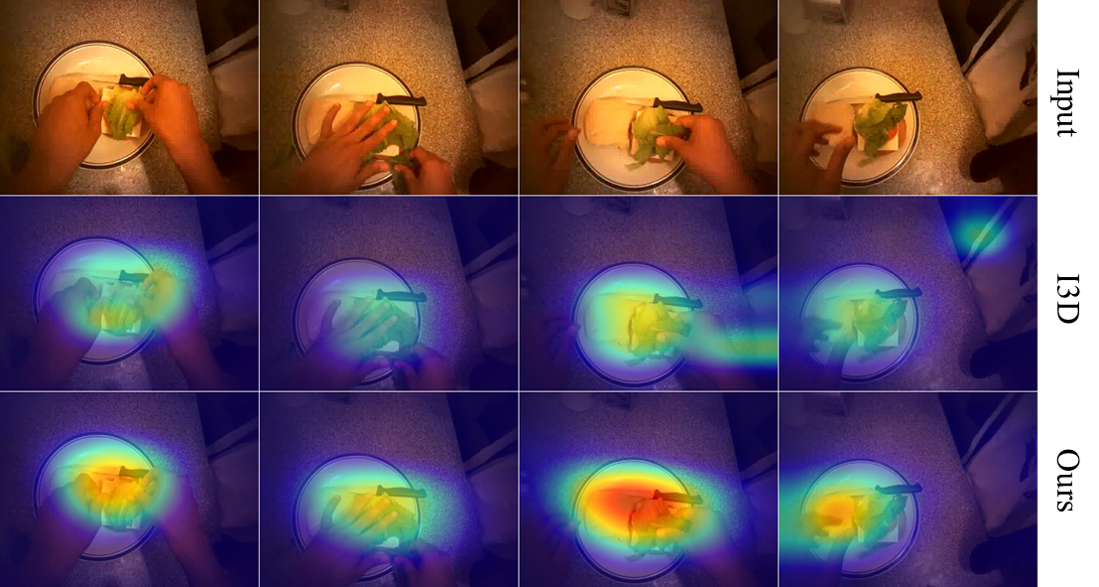}\\[-0.5ex] \hspace*{12.2em}(d)\\[-0.8ex]
    \hspace*{0.14em}\includegraphics[width=0.925\textwidth]{ex-t.png}
  \end{minipage}}
  \caption{Qualitative results of our model and the baseline network (I3D). We use Grad-CAM++~\cite{chattopadhay2018grad} to visualize the spatiotemporal responses of the last layer of each models. We can observe that our method makes the network better at attending objects or regions which are related to the activity. Activity label of (a): ``Move Around bacon'', (b): ``Cut cucumber'', (c): ``Cut bell\_pepper'', (d): ``Put lettuce''.}
  \label{fig:qual}
\end{figure*}

\begin{table*}[t]
\centering
\begin{tabular}{L{6.2cm}|C{1.9cm}C{1.9cm}|C{1.9cm}|C{1.9cm}}
\toprule
\multicolumn{1}{l|}{\multirow{2}{*}{Method}} & \multicolumn{2}{c|}{Using gaze data during} & \multirow{2}{*}{Acc (\%)} & \multirow{2}{*}{Acc$^{*}$ (\%)} \\ & Training & Testing & & \\ \midrule
I3D w/ Gaze & $\checkmark$ & $\checkmark$ & 59.56 & 67.46 \\
I3D w/ Gumbel-Softmax~\cite{jang2016categorical,maddison2016concrete} & $\checkmark$ & & 61.24 & 68.69 \\ \midrule
Ours & $\checkmark$ & & \textbf{62.84} & \textbf{69.58} \\
\bottomrule
\end{tabular}
\caption{Performance comparison of different ablative settings. Interestingly, I3D w/ Gaze that uses gaze data also in the testing process performs the worst. The results demonstrate that our structured gaze modeling with direct optimization is effective in improving the performance of egocentric activity recognition. Qualitative analysis regarding this ablation study is provided in the next section.}
\label{tab:ablation}
\end{table*}

\subsection{Qualitative analysis} \label{subsec:qualitative}
We visualize the response of the last convolutional layer of our model and of I3D~\cite{carreira2017quo} to see how the gaze integration affects the top-down attention of the two networks. We use Grad-CAM++~\cite{chattopadhay2018grad}, which is a recently proposed visualization method for CNNs. It is an improved and generalized version of famous Grad-CAM~\cite{selvaraju2017grad}. It is recently shown that Grad-CAM++ is effective in understanding 3D CNNs on the task of activity recognition by visualizing the attended locations by the networks across space and time. The visualization results are illustrated in Figure~\ref{fig:qual}. We can clearly observe that our model is better at attending activity-related objects or regions. Specifically, our model is more sensitive to the target objects. The baseline network is sometimes distracted by the background objects. The results qualitatively demonstrate that modeling gaze distributions improves the attentional ability of the networks and the performance of egocentric activity recognition.

\subsection{Ablation study}

\begin{figure*}[t]
\centering
\hspace{-6pt}
  \adjustbox{valign=t}{\begin{minipage}[t]{0.49\textwidth}
  \small
    \includegraphics[width=1\textwidth]{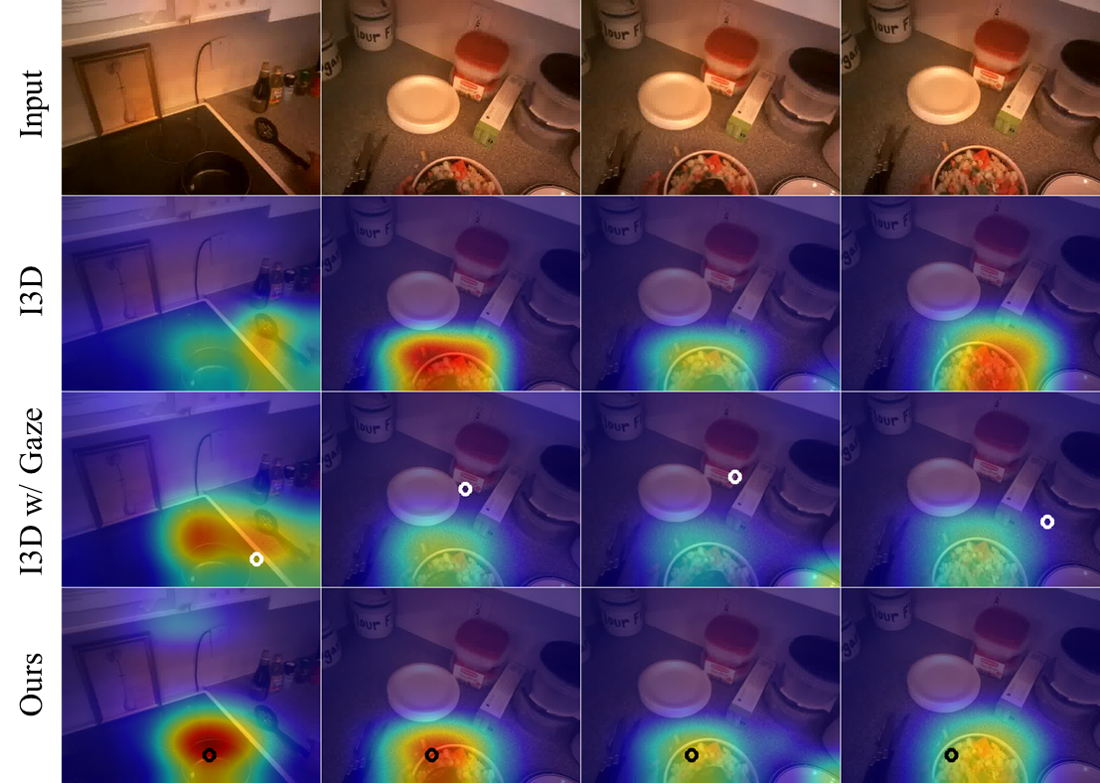}\\[-0.5ex] \hspace*{13.7em}(a)\\[-0.8ex]
    \hspace*{1.54em}\includegraphics[width=0.925\textwidth]{ex-t.png}
  \end{minipage}}
  \hspace{0.029cm}\vline\hspace{0.02cm}
  \adjustbox{valign=t}{\begin{minipage}[t]{0.49\textwidth}
  \small
    \includegraphics[width=1\textwidth]{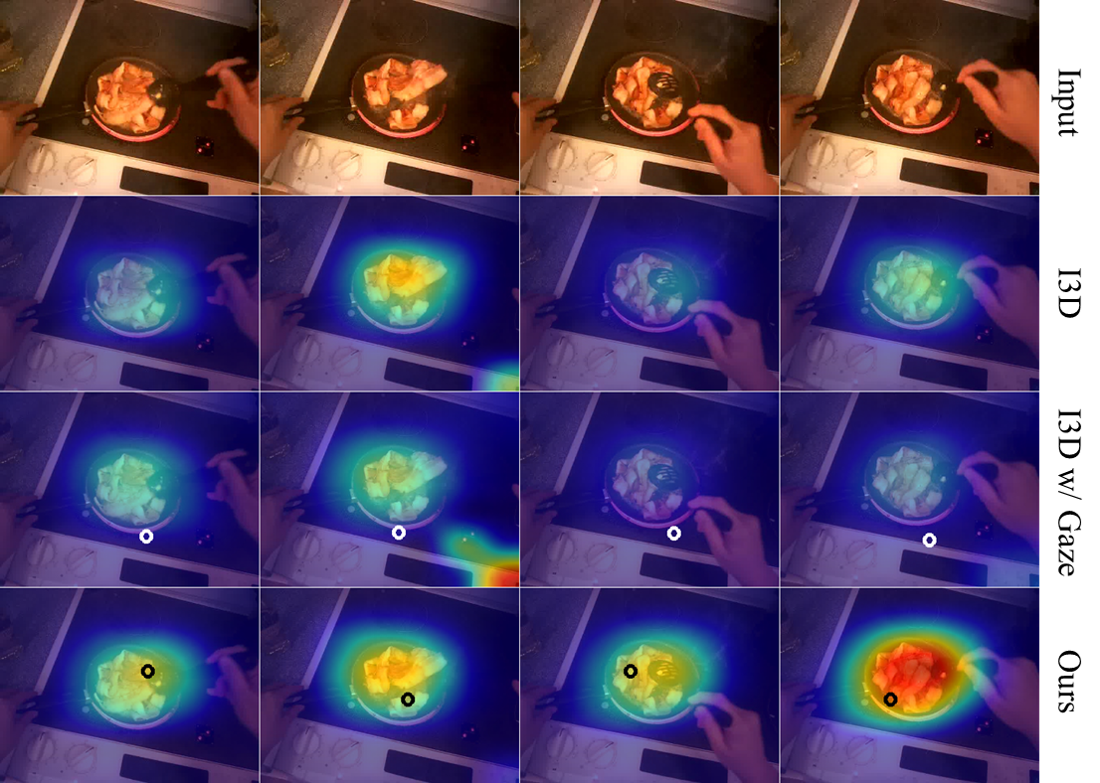}\\[-0.5ex] \hspace*{12.2em}(b)\\[-0.8ex]
    \hspace*{0.04em}\includegraphics[width=0.925\textwidth]{ex-t.png}
  \end{minipage}}
  \caption{Our method is robust to situations where the ground-truth gaze fixations do not carry activity-related information and are misleading. White marks denote ground-truth annotations of gaze fixations and black marks denote the predicted gaze locations. The predicted gaze locations are successfully fixated on the target objects when the ground-truth annotations are misleading. It demonstrates that our structured gaze modeling with direct optimization is effective. Activity label of (a) is ``Mix pasta'' and (b) is ``Move Around bacon''.}
  \label{fig:add}
\end{figure*}

\label{subsec:ablation}
We perform an ablation study on EGTEA dataset~\cite{li2018eye} as reported in Table~\ref{tab:ablation}. ``I3D w/ Gaze'' refers to the method of using the ground-truth gaze annotations without any gaze modeling. For each input segment, the 3D tensor representing the ground-truth gaze locations $\varz_{\text{GT}}$ is first down-sampled to have $3 \times 7 \times7$ dimensions and is applied with a fully-connected layer and the sigmoid function to produce a soft-attention map. This method requires using the gaze data in testing because it does not model the distribution of gaze points. ``I3D w/ Gumbel-Softmax~\cite{jang2016categorical,maddison2016concrete}'' uses the Gumbel-Softmax reparameterization trick to relax the discrete objective to make it continuous. Specifically, it draws a relaxed gaze sample $\varz_{\text{GS}}^{*}$ instead of $\varz^{*}$ in Equation~\ref{eq:opt1} using the following equation: $\varz_{\text{GS}}^{*}=\softmax \big\{ \big(h_{\phi}(\varx,\varz) +\gamma(\varz)\big)/\tau \big\}$. We set $\tau=2$ following the previous work, Li~et~al.~\cite{li2018eye}, that uses the Gumbel-Softmax objective (but takes different gaze modeling approach). The results indicate that our structured gaze modeling with direct optimization is more effective than the other two methods. Interestingly, ``I3D w/ Gaze'' that uses gaze data also in the testing process performs the worst. This is probably because some of the ground-truth gaze annotations are not correlated with the actual visual attention. As mentioned in the introduction, measurement error and other uncertainties (saccadic suppression~\cite{krekelberg2010saccadic} and dissociation~\cite{brefczynski1999physiological,juan2004dissociation}) make the annotated gaze points uninformative and sometimes misleading. We argue that our method is capable of learning only the informative gaze distribution that is related to the activities. We qualitatively analyze these interesting results in the following section.

\subsection{Robustness to misleading gaze fixations}

We perform an additional qualitative analysis to show the robustness of our method to the misleading gaze fixations. Here, misleading gaze points refer to the ground-truth gaze annotations that are not correlated with the actual visual attention. We compare our model with I3D~\cite{carreira2017quo} (without any gaze incorporation) and ``I3D w/ Gaze'' which uses gaze data in training and testing without gaze modeling. We again use Grad-CAM++~\cite{chattopadhay2018grad} to visualize the spatiotemporal activation maps of the last convolutional layer of each model. Figure~\ref{fig:add} illustrates the situations where the ground-truth gaze points are not fixated at the activity-related objects or regions. In these examples, the gaze points are not informative and misleading: the ground-truth gaze points are fixated on the background, not on the pan. This leads to blurry and noisy activation maps of ``I3D w/ Gaze'' because it uses the misleading ground-truth gaze points directly as a soft-attention map. We can observe that our method is robust to such misleading gaze points while ``I3D w/ Gaze'' is not. Specifically, the predicted gaze locations (denoted as black marks) are successfully fixated on the target objects when the ground-truth annotations (denoted as white marks) are not. It demonstrates the effectiveness of our proposed structured gaze modeling with direct optimization.

\section{Additional analysis}
\begin{figure*}[htbp]
\centering
\hspace{-6pt}
  \adjustbox{valign=t}{\begin{minipage}[t]{0.45\linewidth}
  \small
    \includegraphics[width=1\linewidth]{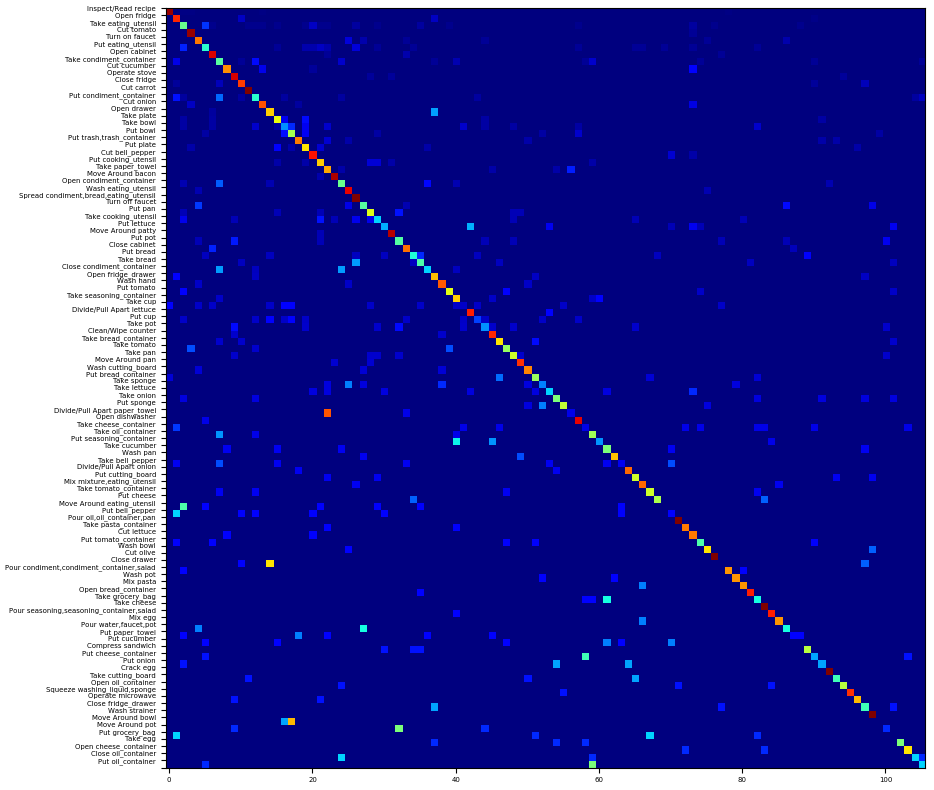}\\[-0.5ex] \hspace*{12.1em}(a) Baseline
  \end{minipage}}
  \hspace{0.6cm}
  \adjustbox{valign=t}{\begin{minipage}[t]{0.45\linewidth}
  \small
    \includegraphics[width=1\linewidth]{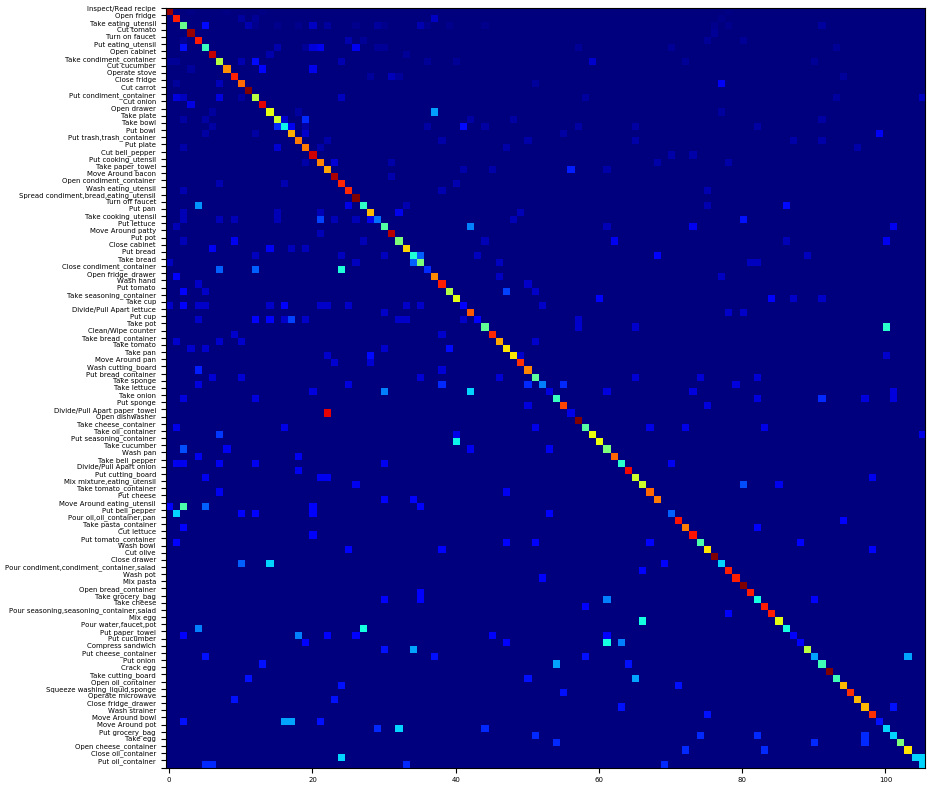}\\[-0.5ex] \hspace*{12.7em}(b) Ours
  \end{minipage}}
  \caption{Confusion matrices for the baseline (I3D~\cite{carreira2017quo}) and ours on the EGTEA dataset~\cite{li2018eye}.}
  \label{fig:conf}
\end{figure*}

\begin{figure*}[htbp]
\centering
\hspace{-6pt}
  \adjustbox{valign=t}{\begin{minipage}[t]{0.45\linewidth}
  \small
    \includegraphics[width=\linewidth]{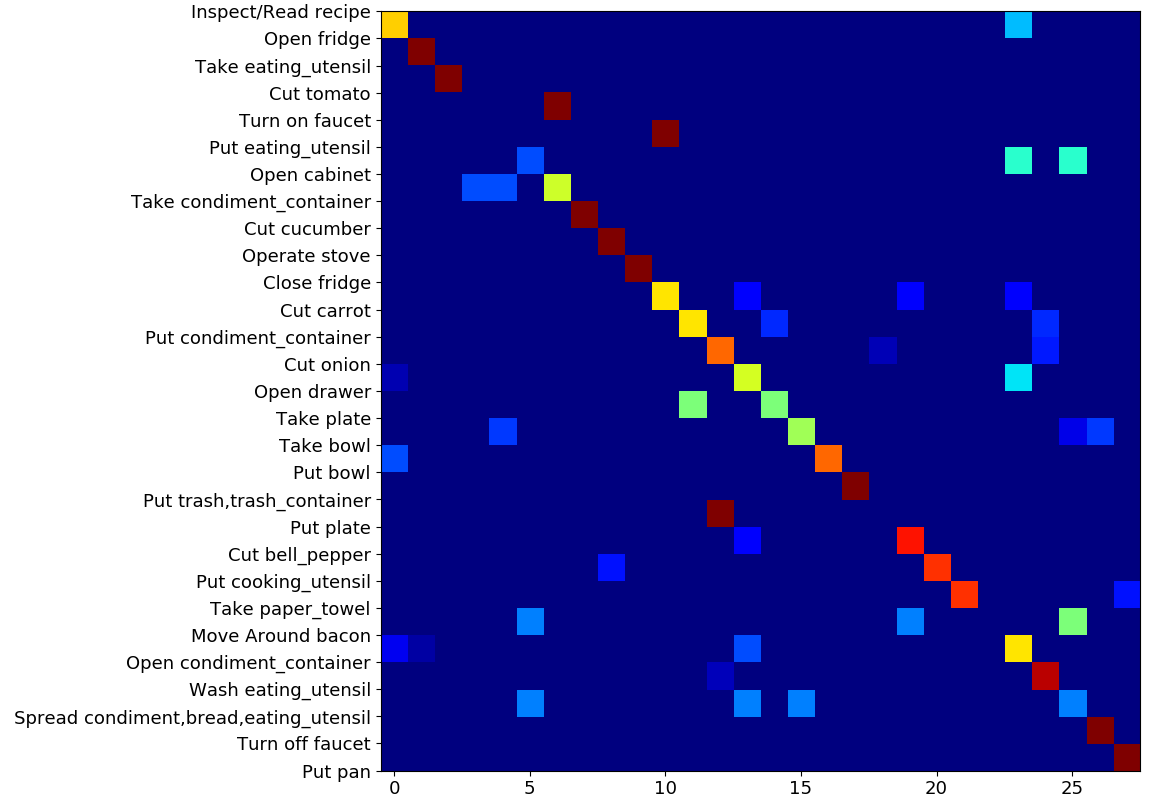}\\[-0.5ex] \hspace*{13.6em}(a) Baseline
  \end{minipage}}
  \hspace{0.6cm}
  \adjustbox{valign=t}{\begin{minipage}[t]{0.45\linewidth}
  \small
    \includegraphics[width=\linewidth]{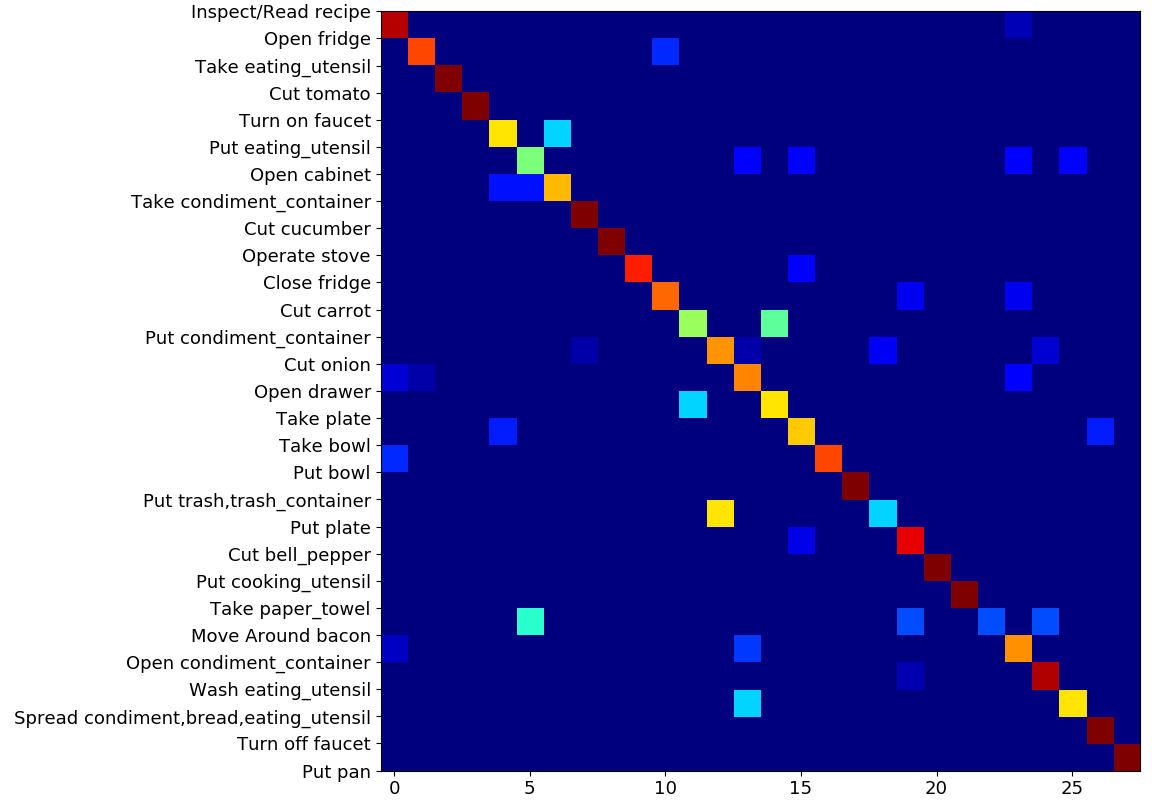}\\[-0.5ex] \hspace*{14.5em}(b) Ours
  \end{minipage}}
  \caption{Confusion matrices for the baseline and ours on 28 classes where our method beats the baseline by a meaningful margin (0.1\%). We can observe that many activities containing ``Cut", ``Take", and ``Put" are better recognized by our gaze incorporated model.}
  \label{fig:conf2}
\end{figure*}

We visualize confusion matrices for the baseline network (I3D~\cite{carreira2017quo}) and our method on the EGTEA dataset~\cite{li2018eye} in Figure~\ref{fig:conf}. Our method outperforms the baseline at least by 0.1\% on 28 classes. For better comparison, we also visualize confusion matrices of the two methods on these 28 classes in Figure~\ref{fig:conf2}. We can observe that many activities containing ``Cut", ``Take", and ``Put" are benefitted from our gaze incorporation.

\section{Conclusion} \label{sec:con}
We have presented an effective method of integrating human gaze into attention on the task of egocentric activity recognition. Incorporating gaze data is non-trivial because there is always uncertainty in the process of recording and the regions near the gaze fixation points are sometimes uninformative. Our method addresses both problems with a probabilistic modeling and an efficient optimization technique. We implement the overall network structures with a simple and powerful 3D CNNs. We evaluate our method in various ways on large-scale datasets. An ablation study demonstrates that incorporating gaze data improves the recognition performance. This is because gaze is correlated with egocentric activity. Moreover, it shows that our proposed structured gaze modeling provides performance improvements by extracting only the informative cues. Interestingly, modeling gaze distribution is more effective in improving the performance than when using ground-truth gaze measurements. We argue that our model is capable of learning only the informative gaze distribution, which is related to the activities of interest. We also qualitatively analyze the effectiveness of our model using the state-of-the-art visualization technique. Our method outperforms all the other previous methods on the task of egocentric activity recognition.


\noindent \textbf{Acknowledgement }
We thank Ryan Szeto and Christina Jung for their valuable comments. This research was, in part, supported by NIST grant 60NANB17D191.



{\small
\bibliographystyle{ieee_fullname}
\bibliography{ref}
}

\end{document}